\documentclass{article}

\PassOptionsToPackage{numbers, compress}{natbib}

\usepackage[final]{neurips_2020}

\usepackage[utf8]{inputenc} 
\usepackage[T1]{fontenc}    
\usepackage{hyperref}       
\usepackage{url}            
\usepackage{booktabs}       
\usepackage{amsfonts}       
\usepackage{nicefrac}       
\usepackage{microtype}      

\usepackage{graphicx}
\usepackage{wrapfig}
\usepackage{subcaption}
\graphicspath{ {./images/} }

\usepackage{tabularx}

\usepackage{amsmath}
\usepackage{gensymb}
\usepackage{bbm}

\title{SelfMatch: Combining Contrastive Self-Supervision and Consistency for Semi-Supervised Learning}

\author{%
  Byoungjip Kim, Jinho Choo, Yeong-Dae Kwon, Seongho Joe, Seungjai Min, Youngjune Gwon\\
  Samsung SDS \\
  \texttt{\{bjip.kim, jinho12.choo, y.d.kwon, drizzle.cho,} \\
  \texttt{seungjai.min, gyj.gwon\}@samsung.com} \\
}

\begin{document}

\maketitle

\begin{abstract}
This paper introduces SelfMatch, a semi-supervised learning method that combines the power of contrastive self-supervised learning and consistency regularization. SelfMatch consists of two stages: (1) self-supervised pre-training based on contrastive learning and (2) semi-supervised fine-tuning based on augmentation consistency regularization. We empirically demonstrate that SelfMatch achieves the state-of-the-art results on standard benchmark datasets such as CIFAR-10 and SVHN. For example, for CIFAR-10 with 40 labeled examples, SelfMatch achieves 93.19\% accuracy that outperforms the strong previous methods such as MixMatch (52.46\%), UDA (70.95\%), ReMixMatch (80.9\%), and FixMatch (86.19\%). We note that SelfMatch can close the gap between supervised learning (95.87\%) and semi-supervised learning (93.19\%) by using only a few labels for each class.  
\end{abstract}

\section{Introduction}
\label{section:introduction}

\begin{wrapfigure}{r}{0.4\textwidth}
  \centering
  \includegraphics[width=0.4\textwidth]{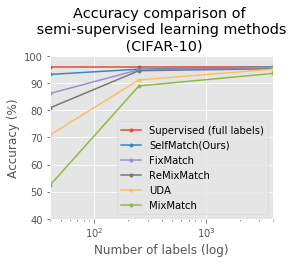}
  \caption{Accuracy comparison of semi-supervised learning methods. SelfMatch can close the gap between supervised and semi-supervised learning by using only a few labels.}
  \label{fig:selfmatch-accuracy-cifar10}
\end{wrapfigure}

Deep neural networks have shown that they can achieve human-level performance in many tasks \cite{he2016deep}\cite{mask2017kaiming}. However, such high performance is usually achieved in a supervised learning setting that exploits a very large number of labeled training examples. Since labeling a large number of examples requires considerable time and cost, label efficient learning algorithms are in high demand.

Semi-supervised learning \cite{chapelle2006semi} is one of the attractive approaches that addresses the label inefficiency problem. Semi-supervised learning enables a deep neural network to learn with small labeled data by leveraging large unlabeled data \cite{lee2013pseudo}\cite{tarvainen2017mean}\cite{laine2016temporal}\cite{verma2019interpolation}\cite{berthelot2019mixmatch}\cite{xie2020unsupervised}\cite{berthelot2020remixmatch}\cite{sohn2020fixmatch}.

Recently, self-supervised learning \cite{kolesnikov2019revisiting} is attracting high attention as a means of unsupervised representation learning. It learns representations by performing a pretext task in which labels can be automatically generated \cite{doersch2015unsupervised}\cite{zhang2016colorful}\cite{noroozi2016unsupervised}\cite{gidaris2018unsupervised}. Also, contrastive self-supervised learning \cite{hjelm2019learning}\cite{oord2018representation}\cite{bachman2019learning}\cite{chen2020simple}\cite{tian2020contrastive}\cite{he2020momentum}\cite{tschannen2020mutual} learns representations by contrasting different views of examples in a task-agnostic way. It is shown that self-supervised learning provides representations good for downstream tasks.

In this paper, we introduce SelfMatch, a semi-supervised learning method that combines the power of contrastive self-supervised learning and consistency regularization. SelfMatch consists of two stages: (1) self-supervised pre-training based on contrastive learning and (2) semi-supervised fine-tuning based on augmentation consistency regularization. We adopt SimCLR \cite{chen2020simple} for self-supervised pre-training, and FixMatch \cite{sohn2020fixmatch} for semi-supervised fine-tuning. SelfMatch achieves new state-of-the-art results on standard benchmarks such as CIFAR-10 \cite{krizhevsky2009learning} and SVHN \cite{netzer2011reading} (see Figure \ref{fig:selfmatch-accuracy-cifar10}). We conjecture that this is because the two stages are complementary in using unlabeled data. Contrastive self-supervised pre-training learns representations by maximizing the mutual information between different views of the data. Meanwhile, consistency regularization further enhances the learned representations by minimizing the cross entropy between class predictions of augmented data (see Appendix \ref{app:discussion}). By combining two closely related but complementary stages, SelfMatch improves from other methods such as S4L \cite{zhai2019s4l} that also uses self-supervised pre-training (see Appendix \ref{app:comparison-with-related-methods}).

Our contributions can be summarized as follows.
\begin{itemize}
    \item We introduce SelfMatch, a semi-supervised learning method that consists of two stages: (1) self-supervised pre-training based on contrastive learning and (2) semi-supervised fine-tuning based on augmentation consistency regularization. (see Figure \ref{fig:selfmatch-overview}). 
    \item We empirically demonstrate that SelfMatch achieves the state-of-the-art results on standard benchmarks such as CIFAR-10 and SVHN (see Table \ref{table:comparison-of-accuracy}). Especially, we show that SelfMatch can close the gap between supervised learning and semi-supervised learning by using only a few labels (see Figure \ref{fig:selfmatch-accuracy-cifar10}).
\end{itemize}
\section{Related Work}
\label{section:related-work}

\subsection{Self-supervised representation learning}

Self-supervised learning \cite{kolesnikov2019revisiting} is a class of methods to unsupervised representation learning. Specifically, it aims to learn representations that can be used for downstream tasks via pretext tasks in which the training labels are automatically generated. Such pretext tasks include context prediction \cite{doersch2015unsupervised}, image colorization \cite{zhang2016colorful}, Jigsaw puzzle solving \cite{noroozi2016unsupervised}, and rotation degree classification (RotNet) \cite{gidaris2018unsupervised}.

Contrastive self-supervised learning is a task-agnostic approach to unsupervised representation learning. It aims to learn representations by contrasting two views from the same or different samples. Its objective functions are based on InfoMax principle \cite{tschannen2020mutual}. This approach includes Deep InfoMax (DIM) \cite{hjelm2019learning}, CPC \cite{oord2018representation}, Augmented Multiscale DIM \cite{bachman2019learning}, SimCLR \cite{chen2020simple}, CMC \cite{tian2020contrastive}, MoCo \cite{he2020momentum}, etc.

\subsection{Semi-supervised learning}
Semi-supervised learning enables a deep neural network to learn with small labeled data by leveraging large unlabeled data. To leverage unlabeled data, diverse methods like consistency regularization \cite{sajjadi2016regularization}\cite{tarvainen2017mean}\cite{laine2016temporal}, entropy minimization \cite{grandvalet2005semi}, and pseudo-labeling \cite{lee2013pseudo} have been proposed.

In pseudo-labeling \cite{lee2013pseudo}, a model is first trained with a small number of labeled examples, and then it assigns pseudo-labels with high confidence on unlabeled examples. And then, the pseudo-labeled examples are used to further train the model in a supervised manner. However, the vanilla pseudo-labeling method suffers from the over-fitting and noisy labels. 

Consistency regularization has been introduced by $\Pi$-model \cite{sajjadi2016regularization}, and is further developed by many following works such as Mean Teacher \cite{tarvainen2017mean}\cite{laine2016temporal}. In the basic consistency regularization, an input image is transformed by a stochastic transformation function and the training objective is to minimize the distance between the model predictions of randomly transformed images by using $L_2$ distance. 

Very recently, advanced consistency regularization methods have been introduced. These methods enhance basic consistency regularization methods with pseudo-labeling and improved data augmentation, and provide very high accuracy that is comparable to supervised learning with full labels. They include ICT \cite{verma2019interpolation}, MixMatch \cite{berthelot2019mixmatch}, UDA \cite{xie2020unsupervised}, ReMixMatch \cite{berthelot2020remixmatch}, and FixMatch \cite{sohn2020fixmatch}.
\section{Method}
\label{section:method}

\begin{figure}[t]
  \centering
  \includegraphics[width=0.85\textwidth]{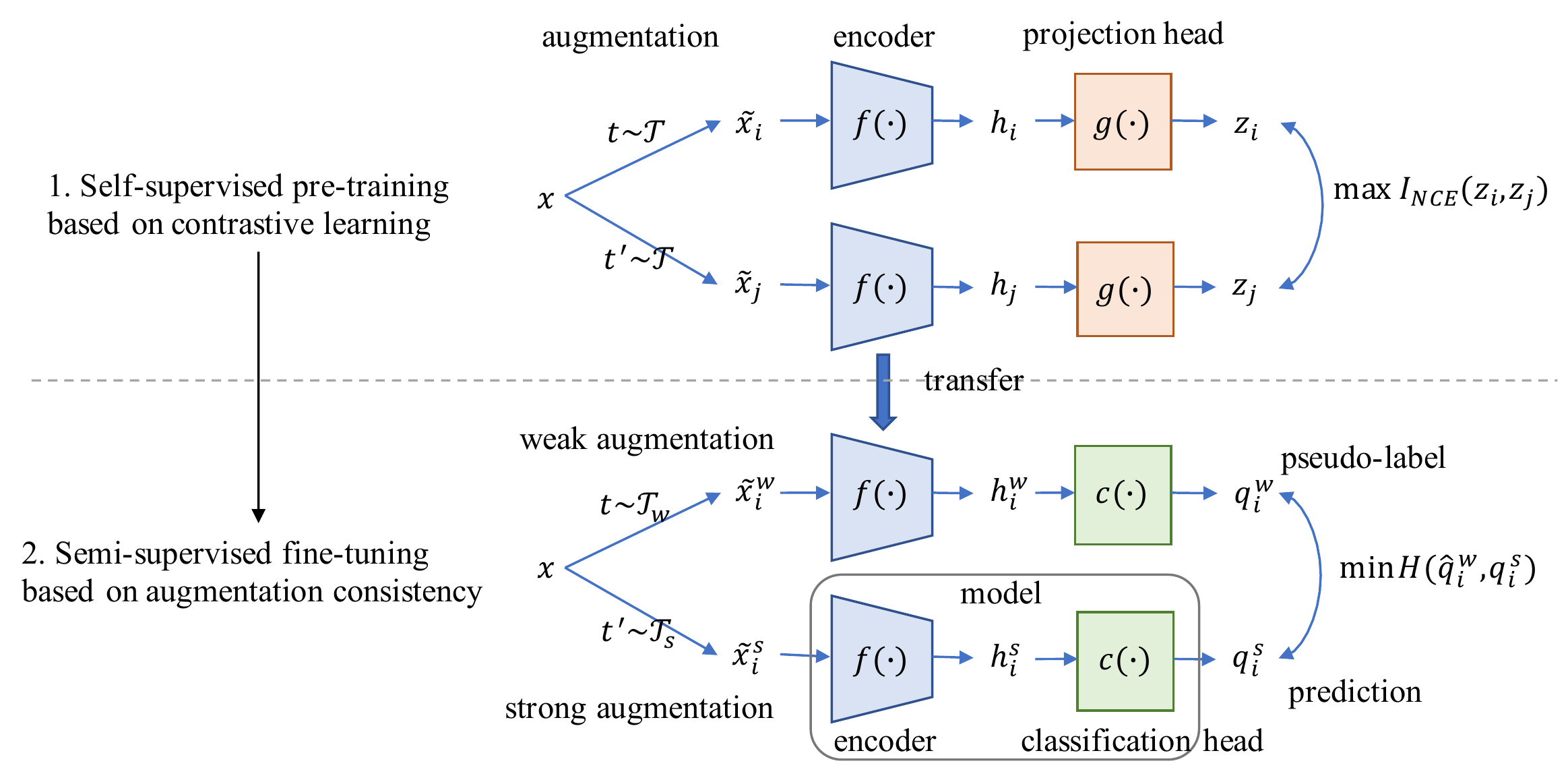}
  \caption{SelfMatch overview.}
  \label{fig:selfmatch-overview}
\end{figure}

An overview of SelfMatch is shown in Figure \ref{fig:selfmatch-overview}. SelfMatch consists of two stages: (1) self-supervised pre-training based on contrastive learning and (2) semi-supervised fine-tuning based on augmentation consistency regularization. The model $p_{model}(y|x)$ consists of an encoder $f(\cdot)$ and a head $c(\cdot)$. 

\subsection{Self-supervised pre-training based on contrastive learning}

To achieve the high accuracy with only a few labels, SelfMatch pre-trains encoder $f(\cdot)$ by leveraging unlabeled data. For this unsupervised pre-training, SelfMatch adopts SimCLR \cite{chen2020simple}, one of the most promising self-supervised learning methods. SimCLR learns representations by encouraging two views $\tilde{x}_i$ and $\tilde{x}_j$ from the same image $x$ to be similar, and two views $\tilde{x}_i$ and $\tilde{x}_k$ $(k \neq i)$ from different images to be dissimilar.

As shown in Figure \ref{fig:selfmatch-overview}, SimCLR consists of four components: data augmentation $\mathcal{T}(\cdot)$, base encoder $f(\cdot)$, projection head $g(\cdot)$, and contrastive loss $\mathcal{L}_c$. For data augmentation, SelfMatch uses random crop and  color distortion. For the base encoder, we use ResNet-34 \cite{he2016deep}. For the projection head, we use a MLP consisting of two layers with Dropout and ReLU activation. Finally, the contrastive loss is formulated as follows: $\mathcal{L}_c = -\log \frac{\textrm{exp}(\textrm{sim}(z_i,z_j)/\tau)}{\sum_{k=1}^{2B} \mathbbm{1}(k \neq i)\textrm{exp}(\textrm{sim}(z_i,z_k)/\tau)}$, where $\textrm{sim}(\cdot)$ is a similarity measure function, $\tau$ is a temperature parameter scaling the similarity, $\mathbbm{1}(k \neq i)$ is an indicator function evaluating to 1 iff $k \neq i$, and $B$ is a batch size. We set hyperparameters to $\tau = 0.5$ and $B = 512$.

\subsection{Semi-supervised fine-tuning based on augmentation consistency regularization}

Unlike the related methods such as S4L \cite{zhai2019s4l} (using semi-supervised fine-tuning based on VAT \cite{miyato2018virtual}) and SimCLRv2 \cite{chen2020big} (using supervised fine-tuning with few labels), SelfMatch exploits semi-supervised fine-tuning based on augmentation consistency. For augmentation consistency, we adopt FixMatch \cite{sohn2020fixmatch}. It encourages the consistent prediction between weakly and strongly augmented examples ($\tilde{x}_i^w$ and $\tilde{x}_i^s$). More concretely, FixMatch uses a model output $q_i^w$ of a weakly augmented input as the pseudo-label for a strongly augmented input $\tilde{x}_i^s$ (see Figure \ref{fig:selfmatch-overview}). For weak augmentation, we use random crop and random horizontal flip. For strong augmentation, SelfMatch adopts RandAugment (RA) \cite{cubuk2020randaugment}, an effective automated augmentation method (see Appendix \ref{app:randaugment}). For the classification head $c(\cdot)$, SelfMatch uses a MLP consisting of two layers with Dropout \cite{srivastava2014dropout} and ReLU activation.  

The loss function of FixMatch consists of supervised and unsupervised loss: $\mathcal{L}_{semi} = \mathcal{L}_s + \lambda_u \mathcal{L}_u$. The supervised loss $\mathcal{L}_s$ is formulated as follows: $\mathcal{L}_s = \frac{1}{B} \sum_{i=1}^{B} H(p_i, q_i^w)$, where $B$ is a batch size, $p_i$ is a one-hot encoded label of sample $x_i$, and $q_i^w$ is a model output $p_{model}(y|\tilde{x}_i^w)$ of a weakly augmented input $\tilde{x}_i^w$. For the experiments, we set $B$ to 64.

The unsupervised loss $\mathcal{L}_u$ is formulated as follow: $\mathcal{L}_u = \frac{1}{\mu B} \sum_{i=1}^{\mu B} \mathbbm{1}(\textrm{max}(q_{i}^{w}) \geq c) H(\hat{q}_i^w, q_i^s)$, where $\mathbbm{1}(\textrm{max}(q_{i}^{w}) \geq c)$ is an indicator function, $c$ is a confidence threshold, $B$ is a batch size, and $\mu$ is the ratio of labeled and unlabeled samples in a batch. Here, $\hat{q}_i^w$ is $\arg \max (q_i^w)$ and $q_i^s$ is a model output of a strongly augmented input $\tilde{x}_i^s$. We set hyperparameters to $c$ = 0.95, $\mu$ = 7, and $\lambda_u$ = 1.

For training, SelfMatch uses a standard SGD optimizer with momentum $\beta = 0.9$ and exploits cosine learning rate decay \cite{loshchilov2017sgdr} with the initial learning rate $\eta = 0.03$. Also, it utilizes Exponential Moving Average (EMA) \cite{tarvainen2017mean} with weight decay 0.999 for stable training and inference.
\section{Experiments}
\label{section:experiments}

We evaluate the performance of SelfMatch on standard classification benchmark datasets such as CIFAR-10 \cite{krizhevsky2009learning} and SVHN \cite{netzer2011reading}. To see the label efficiency, we perform evaluation with varying amount of labeled data. For CIFAR-10, we use 50,000 unlabeled training images and 10,000 testing images. For SVHN, we use 73,357 unlabeled training images and 26,032 test images. We present SelfMatch results using the average value from three evaluation runs with different random seeds.

\begin{table}[t]
  \caption{Comparison of accuracy for CIFAR-10 and SVHN.}
  \label{table:comparison-of-accuracy}
  
  \centering
  \begin{tabular}{m{2.1cm} m{1.5cm}m{1.5cm}m{1.6cm} m{1.5cm}m{1.5cm}m{1.6cm}}
    \toprule
     & \multicolumn{3}{c}{CIFAR-10} & \multicolumn{3}{c}{SVHN} \\
    
    \cmidrule(lr){2-4} \cmidrule(lr){5-7}
    \noalign{\smallskip}
    \textbf{Method}             & 40 labels & 250 labels & 4000 labels & 40 labels & 250 labels & 1000 labels \\
  
    \cmidrule(lr){1-1} \cmidrule(lr){2-4} \cmidrule(lr){5-7}
    \noalign{\smallskip}
    Supervised                  & \multicolumn{3}{c}{\textbf{95.87}} & \multicolumn{3}{c}{\textbf{97.41}} \\
    
    \cmidrule(lr){1-1} \cmidrule(lr){2-4} \cmidrule(lr){5-7}
    \noalign{\smallskip}
    Pseudo-Label                & - & 50.22$\pm$0.43 & 83.91$\pm$0.28 & - & 79.79$\pm$1.09 & 90.06$\pm$0.61 \\
    
    \cmidrule(lr){1-1} \cmidrule(lr){2-4} \cmidrule(lr){5-7}
    \noalign{\smallskip}
    $\Pi$-Model                 & - & 45.74$\pm$3.87 & 85.99$\pm$0.38 & - & 81.04$\pm$1.92 & 92.46$\pm$0.36 \\
    Mean Teacher                & - & 67.68$\pm$2.30 & 90.81$\pm$0.19 & - & 96.43$\pm$0.11 & 96.58$\pm$0.07 \\
    
    \cmidrule(lr){1-1} \cmidrule(lr){2-4} \cmidrule(lr){5-7}
    \noalign{\smallskip}
    MixMatch		            & 52.46$\pm$11.50 & 88.95$\pm$0.86 & 93.58$\pm$0.10 & 57.45$\pm$14.53 & 96.02$\pm$0.23 & 96.5$\pm$0.28 \\
    UDA                         & 70.95$\pm$5.93 & 91.18$\pm$1.08 & 95.12$\pm$0.18 & 47.37$\pm$20.51 & 94.31$\pm$2.76 & 97.54$\pm$0.24 \\
    ReMixMatch                  & 80.90$\pm$9.64 & 94.56$\pm$0.05 & 95.28$\pm$0.13 & \textbf{96.66}$\pm$0.20 & 97.08$\pm$0.48 & 97.35$\pm$0.08 \\
    FixMatch(RA)     	        & 86.19$\pm$3.37 & 94.93$\pm$0.65 & 95.74$\pm$0.05 & 96.04$\pm$2.17 & \textbf{97.52}$\pm$0.38 & \textbf{97.72}$\pm$0.11 \\ 
    
    \cmidrule(lr){1-1} \cmidrule(lr){2-4} \cmidrule(lr){5-7}
    \noalign{\smallskip}
    \textbf{SelfMatch} & \textbf{93.19}$\pm$1.08 & \textbf{95.13}$\pm$0.26 & \textbf{95.94}$\pm$0.08 & \textbf{96.58}$\pm$1.02 & \textbf{97.37}$\pm$0.43 & \textbf{97.49}$\pm$0.07 \\
    
    \bottomrule
  \end{tabular}
\end{table}

\setlength{\tabcolsep}{7pt}
\begin{table}[t]
  \caption{Effect of self-supervised pre-training on semi-supervised learning (CIFAR-10).}
  \label{table:ablation-study}
  
\centering
  \begin{tabular}{m{1.5cm} m{0.8cm}m{0.8cm}m{0.8cm} m{1.5cm} m{0.8cm}m{0.8cm}m{0.8cm}}
    \toprule
    \textbf{Method}  & 40 & 250 & 4000 & \textbf{Method} & 40 & 250 & 4000 \\
    
    \cmidrule(lr){1-1} \cmidrule(lr){2-4} \cmidrule(lr){5-5} \cmidrule(lr){6-8}
    MixMatch & 52.46 & 88.95 & 93.58 & FixMatch & 86.19 & 94.93 & 95.74\\
    
    \cmidrule(lr){1-1} \cmidrule(lr){2-4} \cmidrule(lr){5-5} \cmidrule(lr){6-8}
    SimCLR + MixMatch & 63.42 (+10.96) & 92.58 (+3.63) & 94.96 (+1.38) & SimCLR + FixMatch & 93.19 (+7.00) & 95.13 (+0.2) & 95.94 (+0.2) \\
    
    \bottomrule
  \end{tabular}
\end{table}

\paragraph{CIFAR-10}
Table ~\ref{table:comparison-of-accuracy} shows the results for CIFAR-10. SelfMatch achieves the state-of-the-art for all number of labeled examples. More important, at 40 labeled examples (4 labels for each class), SelfMatch achieves 93.19\% accuracy that outperforms the strong previous methods including MixMatch \cite{berthelot2019mixmatch} (52.46\%), UDA \cite{xie2020unsupervised} (70.95\%), ReMixMatch \cite{berthelot2020remixmatch} (80.9\%), and FixMatch \cite{sohn2020fixmatch} (86.19\%). It is remarkable that the gap  between the supervised learning (95.87\%) and the semi-supervised learning (93.19\%) can be narrowed down to a few percent with only 4 labels per class (see also Figure \ref{fig:selfmatch-accuracy-cifar10}). To further demonstrate the effectiveness of SelfMatch, a comparison of learning curves of different methods is presented in Appendix \ref{app:comparison-of-learning-curves}.

\paragraph{SVHN}
Table ~\ref{table:comparison-of-accuracy} shows the results for SVHN. SelfMatch achieves very high accuracy that are within the margin of error of other state-of-the-art results.

\paragraph{Ablation study}
Table ~\ref{table:ablation-study} shows the effect of self-supervised pre-training for two representative semi-supervised learning methods, MixMatch \cite{berthelot2019mixmatch} and FixMatch \cite{sohn2020fixmatch}. At 40 labeled examples, self-supervised pre-training improves MixMatch by 10.96\%, and FixMatch by 7.00\%. Further ablation study can be found in Appendix \ref{app:further-abliation-study}.
\section{Conclusion}
\label{section:conclusion}

In this paper, we introduce SelfMatch, a semi-supervised learning method that consists of two stages: (1) self-supervised pre-training based on contrastive learning and (2) semi-supervised fine-tuning based on augmentation consistency regularization. We empirically demonstrate that SelfMatch can close the gap between supervised learning and semi-supervised learning using only a few labels.

\medskip
\small
\bibliography{selfmatch_bib}

\normalsize
\appendix
\newpage
\section{Comparison with related method}
\label{app:comparison-with-related-methods}

A comparison with a closely related method is shown in Table \ref{table:comparison-with-related-method}. Very recently, Zhai et al. presented S4L \cite{zhai2019s4l}, a semi-supervised learning method that consists of three stages: (1) self-supervised pre-training based on rotation prediction (RotNet \cite{gidaris2018unsupervised}), (2) semi-supervised fine-tuning based on adversarial consistency (VAT \cite{miyato2018virtual}), and (3) supervised fine-tuning with a few labeled data. SelfMatch is different from S4L. First, SelfMatch proposes to use contrastive self-supervised pre-training (e.g., SimCLR \cite{chen2020simple}, MoCo \cite{he2020momentum}, etc). Since contrastive self-supervised learning is task-agnostic, it provides better representations for downstream tasks. Second, SelfMatch proposes to use semi-supervised fine-tuning based on augmentation consistency (e.g., FixMatch \cite{sohn2020fixmatch}, ReMixMatch \cite{berthelot2020remixmatch}, etc.). Since contastive self-supervised learning and augmentation consistency regularization are commonly based on data augmentation, SelfMatch is architecturally simpler and has more rooms to be optimized.

\begin{table}[h]
  \caption{A comparison with a related method.}
  \label{table:comparison-with-related-method}

  \centering
  \begin{tabular}{m{2.5cm} m{4.0cm} m{4.5cm}}
    \toprule
    \textbf{Phase} & \textbf{S4L} \cite{zhai2019s4l} & \textbf{SelfMatch} (Ours) \\
    
    \hline
    \noalign{\smallskip}
    1. Pre-training & Task-specific self-supervised (RotNet \cite{gidaris2018unsupervised}) & Task-agnostic self-supervised (SimCLR \cite{chen2020simple}) \\
    
    \hline
    \noalign{\smallskip}
    2. Fine-tuning A & Semi-supervised (adversarial consistency, VAT \cite{miyato2018virtual}) & Semi-supervised (augmentation consistency, FixMatch \cite{sohn2020fixmatch}) \\
    
    \hline
    \noalign{\smallskip}
    3. Fine-tuning B & Supervised with few labels & - \\
    \bottomrule
  \end{tabular}
\end{table}
\section{Transformations in RandAugment}
\label{app:randaugment}

SelfMatch uses RandAugment \cite{cubuk2020randaugment} for strong augmentation in the semi-supervised fine-tuning stage. Table \ref{table:randaugment} shows the list of transformations used in RandAugment. At each strong data augmentation process, two transformations with random magnitude are randomly selected and serially performed.

\begin{table}[h]
    \caption{List of transformations used in RandAugment \cite{cubuk2020randaugment}.}
    \label{table:randaugment}
    
    \centering
    \begin{tabular}{m{2.2cm} m{9.0cm}}
        \toprule
        \textbf{Transformation} & \textbf{Description} \\
        \hline
        \noalign{\smallskip}
        Autocontrast   & Maximizing the image contrast. \\
        Brightness     & Adjusting the brightness of the image. (0: black image, 1: original image) \\
        Color          & Adjusting the color balance of the image. (0: black \& while image, 1: original image) \\
        Contrast       & Adjusting the contrast of the image. (0: a gray image, 1: original image) \\
        Equalize       & Equalizing the image histogram. \\
        Identity       & The original image. \\
        Posterize      & Reducing each pixel to [4, 8] bits. \\
        Rotate         & Rotating the image by [-30, 30] degrees. \\
        Sharpness      & Adjusting the sharpness of the image. (0: blurred image, 1: original image) \\
        Shear-x & Shearing the image along the horizontal axis with rate [-0.3, 0.3]. \\
        Shear-y        & Shearing the image along the vertical axis with rate [-0.3, 0.3]. \\
        Solarize       & Inverting all pixels above a threshold value of [0, 1]. \\
        Translate-x    & Translating the image horizontally by ([-0.3, 0.3] x image width) pixels. \\
        Translate-y    & Translating the image vertically by ([-0.3, 0.3] x image height) pixels. \\
        \bottomrule
    \end{tabular}
\end{table}
\newpage
\section{Further ablation study}
\label{app:further-abliation-study}

Table \ref{table:further-ablation-study} shows further ablation study. More specifically, it shows baseline accuracy results of self-supervised pre-training for CIFAR-10 \cite{krizhevsky2009learning} with 4000 labels. Usually, in unsupervised representation learning, the performance of learned representation is measured by using a linear classification head and full labeled data \cite{chen2020simple}. As shown in Table \ref{table:further-ablation-study}, when following such an evaluation protocol for CIFAR-10, SimCLR \cite{chen2020simple} provides about 92.27\% accuracy. When using a MLP head consisting of two layers with Dropout \cite{srivastava2014dropout} and ReLU activation, SimCLR provides about 92.69\% accuracy (+0.42\%). In semi-supervised learning setting with 4000 labels, it decreases to 91.36\% (-1.33\%). This can be considered as a baseline method that exploits self-supervised pre-training. More specifically, the baseline method consists of (1) self-supervised pre-training and (2) supervised fine-turning with 4000 labels. Unlike this, using semi-supervised fine-truing based on augmentation consistency (FixMatch \cite{sohn2020fixmatch}) in the second stage, SelfMatch improves the baseline method up to 95.94\% in a large margin (+4.58\%).

\begin{table}[h]
    \caption{Baseline accuracy results of self-supervised pre-training (CIFAR-10).}
    \label{table:further-ablation-study}
    
    \centering
    \begin{tabular}{m{1.5cm} lll l}
        \toprule
        \textbf{Method} & \textbf{Encoder} (ResNet-34) & \textbf{Head} & \textbf{Labels} & \textbf{Accuracy} (\%) \\

        \hline
        \noalign{\smallskip}
        Supervised & random init. & Linear (512 $\times$ 10) & 50000 & 95.87 \\
        Supervised & random init. & Linear (512 $\times$ 10) & 4000 & 79.74 \\
        
        \hline
        \noalign{\smallskip}
        SimCLR & frozen & Linear (512 $\times$ 10) & 50000 & 92.27 \\
        SimCLR & frozen & MLP (2 layers) & 50000 & 92.69 \\
        SimCLR & frozen & Linear (512 $\times$ 10) & 4000 & 90.99 \\
        SimCLR & frozen & MLP (2 layers) & 4000 & \textbf{91.36} \\
        
        \hline
        \noalign{\smallskip}
        SelfMatch & fine-tuned with FixMatch & MLP (2 layers) & 4000 & \textbf{95.94} (+4.58) \\
        
        \bottomrule     
    \end{tabular}
\end{table}
\section{Comparison of learning curves}
\label{app:comparison-of-learning-curves}

\begin{figure}[h]
  \begin{subfigure}{0.3\textwidth}
  \includegraphics[width=0.95\linewidth, height=4cm]{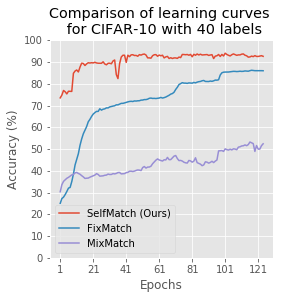}
  \caption{40 labels}
  \label{fig:learning-curve-cifar10-40labels}
  \end{subfigure}
  ~
  \begin{subfigure}{0.3\textwidth}
  \includegraphics[width=0.95\linewidth, height=4cm]{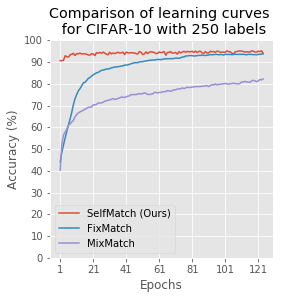}
  \caption{250 labels}
  \label{fig:learning-curve-cifar10-250labels}
  \end{subfigure}
  ~
  \begin{subfigure}{0.3\textwidth}
  \includegraphics[width=0.95\linewidth, height=4cm]{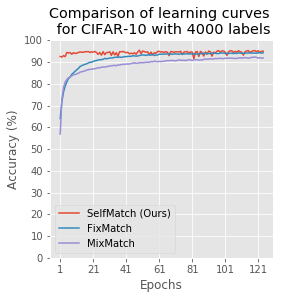}
  \caption{4000 labels}
  \label{fig:learning-curve-cifar10-4000labels}
  \end{subfigure}
  
  \caption{A comparison of learning curves for CIFAR-10.}
  \label{fig:comparison-of-learning-curves}
\end{figure}

To demonstrate the effectiveness of SelfMatch in more detail, we present a comparison of learning curves for CIFAR-10 in Figure \ref{fig:comparison-of-learning-curves}. Each figure shows a comparison of learning curves of three most advanced methods including SelfMatch (ours), FixMatch \cite{sohn2020fixmatch} and MixMatch \cite{berthelot2019mixmatch} for varying number of labels (i.e., 40, 250, 4000 labels). Note that the learning curves of SelfMatch only present the second stage of SelfMatch, that is the semi-supervised fine-tuning stage after the self-supervised pre-training stage. 

As shown in Figure \ref{fig:comparison-of-learning-curves}, compared to other methods, SelfMatch not only achieves higher accuracy, but also reaches such high accuracy more rapidly. The accuracy gain is more prominent in the case where only a few labels are available (see Figure \ref{fig:learning-curve-cifar10-40labels}). These results are also summarized in Table \ref{table:comparison-of-accuracy}. More importantly, SelfMatch reaches such high accuracy more rapidly. For 250 and 4000 labels, SelfMatch reaches an accuracy of greater than 90\% through just one epoch. In the case of 40 labels, SelfMatch achieves an accuracy of 90\% through around 10 epochs, while FixMatch and MixMatch do not reach such high accuracy even though iterating through a very large number of training epochs.       
\newpage
\section{Discussion}
\label{app:discussion}

\paragraph{Relation between contrastive self-supervision and consistency regularization}
In this paper, we introduce SelfMatch, a semi-supervised learning method that combines the power of contrastive self-supervised learning and consistency regularization. To further motivate the reason why we combine these two approaches, we summarize the relation between the two approaches in Table \ref{table:relation-between-contrastive-and-consistency}. Note that the relation is based on SimCLR \cite{chen2020simple} and FixMatch \cite{sohn2020fixmatch} adopted by SelfMatch. If other methods are considered for each approach, the comparison would be slightly different.

SelfMatch consists of two stages: (1) self-supervised pre-training based on contrastive learning and (2) semi-supervised fine-tuning based on consistency regularization. We empirically found that SelfMatch not only achieves higher accuracy, but also reaches such high accuracy more rapidly. As mentioned in Section \ref{section:introduction}, we conjecture that this is because the two stages are complementary in using unlabeled data. Contrastive self-supervised learning methods such as SimCLR \cite{chen2020simple} use loss functions based on the InfoMax principle \cite{linsker1988self}, and the mutual information is usually estimated by InfoNCE \cite{oord2018representation}. Also, Michael et al. \cite{tschannen2020mutual} have shown that InfoNCE losses are related with triplet losses \cite{sohn2016improved}. We have a similar perspective. In the pre-training stage of SelfMatch, contrastive self-supervised learning learns representations by maximizing a lower bound (e.g., InfoNCE) of the mutual information between different views of the data. This InfoNCE loss can be interpreted as encouraging two objectives: (1) maximizing the similarity between different views from the same sample and (2) minimizing the similarity between different views from different samples. The former objective can be considered as using intra-sample statistics, and the later objective can be understood as leveraging inter-sample statistics. Meanwhile, in the fine-tuning stage of SelfMatch, consistency regularization \cite{sohn2020fixmatch} further enhances the learned representations by minimizing the cross entropy (or distance) between class predictions of different views from the same sample. This objective can be considered as exploiting intra-sample statistics. 

Furthermore, two approaches are slightly different in exploiting intra-sample statistics. The target entities to optimize are normalized representations in contrastive self-supervised learning, while they are class probability distributions in consistency regularization. Also, consistency regularization of FixMatch exploits asymmetric augmentation (i.e., weak and strong augmentation of the same sample).

\begin{table}[h]
  \caption{Relation between contrastive self-supervision and consistency regularization.}
  \label{table:relation-between-contrastive-and-consistency}

  \centering
  \begin{tabular}{m{2.0cm} m{5.3cm} m{5.3cm}}
    \toprule
     & \textbf{Contrastive self-supervision} & \textbf{Consistency regularization}\\
     
    \hline
    \noalign{\smallskip}
    Method & SimCLR \cite{chen2020simple} & FixMatch \cite{sohn2020fixmatch} \\
     
    \hline
    \noalign{\smallskip}
    Loss function & $-\log \frac{\textrm{exp}(\textrm{sim}(z_i,z_j)/\tau)}{\sum_{k=1}^{2B} \mathbbm{1}(k \neq i)\textrm{exp}(\textrm{sim}(z_i,z_k)/\tau)}$ & $\frac{1}{\mu B} \sum_{i=1}^{\mu B} \mathbbm{1}(\textrm{max}(q_{i}^{w}) \geq c) H(\hat{q}_i^w, q_i^s)$ \\
    
    \hline
    \noalign{\smallskip}
     & Maximizing the similarity between different views form the same sample ($z_i$ and positive sample $z_j$) & Minimizing the cross entropy (or distance) between predictions of different views of the same sample ($\hat{q}_i^w$ and $q_i^s$) \\
     & Minimizing the similarity between different views from different samples ($z_i$ and negative sample $z_k$) & \\
    
    \hline
    \noalign{\smallskip}
    Negative samples & Yes & No \\
     
    \hline
    \noalign{\smallskip}
    Target entity & Representations ($z_i$) & Class probability distributions ($q_i$)\\
    
    \hline
    \noalign{\smallskip}
    Transformation & Symmetric & Asymmetric \\
     & (two weak augmentations) & (weak and strong augmentations) \\
    
    \bottomrule
  \end{tabular}
\end{table}

\paragraph{Large-scale experiment}
In this paper, we imperially demonstrate the effectiveness of SelfMatch by using standard benchmark datasets such as CIFAR-10 \cite{krizhevsky2009learning} and SVHN \cite{netzer2011reading}. However, these datasets are rather small. Therefore, we are doing experiments by using larger datasets such as STL-10 \cite{coates2011analysis} and ImageNet \cite{deng2009imagenet}, and will add the results in an extended version of this paper.

\end{document}